# On the Winograd Schema: Situating Language Understanding in the Data-Information-Knowledge Continuum

**Walid S. Saba**

Astound.ai, 111 Independence Drive, Menlo Park, CA 94025 USA
walid@astound.ai

**Abstract**

The Winograd Schema (WS) challenge, proposed as an alternative to the Turing Test, has become the new standard for evaluating progress in natural language understanding (NLU). In this paper we will not however be concerned with how this challenge might be addressed. Instead, our aim here is threefold: (*i*) we will first formally 'situate' the WS challenge in the data-information-knowledge continuum, suggesting where in that continuum a good WS resides; (*ii*) we will show that a WS is just a special case of a more general phenomenon in language understanding, namely the *missing text phenomenon* (henceforth, MTP) - in particular, we will argue that what we usually call *thinking* in the process of language understanding involves discovering a significant amount of 'missing text' - text that is not explicitly stated, but is often implicitly assumed as shared background knowledge; and (*iii*) we conclude with a brief discussion on why MTP is inconsistent with the data-driven and machine learning approach to language understanding.

## Introduction

Consider the sentence in (1):

(1) *Dave told everyone in school that he wants to be a guitarist, because he thinks **it** is a great instrument.*

Short of having access to relevant background knowledge, quantitative (statistical, data-driven and machine learning) methods would, and with a high degree of certainty, erroneously resolve "it" in (1) since the correct referent is **not even in the data**, but, as a 5-year old would correctly infer, is an object that is implicitly implied by the semantic and cognitive content of the text: a-guitarist-plays-a-guitar and a-guitar-is-a-musical-instrument. Undoubtedly, it is this kind of *thinking* that Alan Turing had in mind when he posed the question "Can Machines Think?" (Turing, 1950), suggesting further that a machine that intelligently communicates in ordinary spoken language, much like humans do, must be a *thinking* machine[1]. As recently suggested by Levesque et. al. (2012), however, the Turing Test left room for the possibility of some systems to pass the test, not because anything we might call thinking is going on, but by trickery and deception. As Levesque et. al. point out, systems that have participated in the Loebner competition (Shieber 1994) usually use deception and trickery by throwing in "elaborate wordplay, puns, jokes, quotations, clever asides, emotional outbursts," while avoiding clear answers to questions that a 5-year old would be very comfortable in correctly answering. In addressing these shortcomings, Levesque et. al. suggested what they termed the Winograd Schema (WS) challenge, illustrated by the following example[2]:

(2) The **city councilmen** refused **the demonstrators** a permit because they
  a. **feared** violence.
  b. **advocated** violence.

The question posed against this sentence would be: what does "they" refer to in (2a), and what does it refer to in (2b)? The answer seems so obvious to humans that reason using relevant commonsense background knowledge (e.g., demonstrators are more prone to advocate violence than the governing body of a certain city, while the latter are more likely to fear the violence) and thus a machine that correctly resolves such references would be performing what we might call thinking. Levesque points out however that care should be taken in the design of such queries so as to avoid the pitfalls of the original Turing Test, namely that a program should not be able to pass the test by performing simple syntactic level and pattern matching computations. For example, simple word co-occurrence data obtained from a corpus analysis might be all that is needed to make the correct guess in (3), while the same is not true in (4).

---

[1] Although this is not the subject of this paper, we however would like to unequivocally concur that language, that infinite object that is tightly related to our capacity to have an infinite number of thoughts, is the ultimate test for thinking machines. Thus, while several accomplishments in computation are usually attributed to AI, most of these tasks deal with finding a near optimal solution from a finite set of possibilities and are hardly performing what we might call thinking. For example, and although the search space is very large, playing chess is ultimately a matter of scoring more paths than the opponent, thus making the probability of winning in the long run certain. The same can also be said of pattern (sound and image) recognition systems that, essentially, find regularities in data. True human-level scene analysis, beyond lower level image recognition that the most primitive of species can perform, would also require reasoning similar to that required in language understanding. In this regard we believe the recently proposed visual Turing Test (Geman et al., 2014) is a step in the right direction.
[2] The example in (2) was originally discussed by Terry Winograd (1972), after whom the challenge was named.

(3) *The **women** stopped taking **the pills** because they were*
   a. ***pregnant***.
   b. ***carcinogenic***.

(4) *The **trophy** would not fit into **the brown suitcase** because it was too*
   a. ***small***.
   b. ***big***.

Levesque calls the query in (4) "Google-proof", since having access to a large corpus would not help here as the frequencies of the antonyms "small" and "big" in similar contexts should in principle be the same, as studies have indicated (e.g., Kostic, 2017). This is not the same in (3), however, where a purely ***quantitative*** system would pass many queries based on simple co-occurrence data; for example, the likelihood of carcinogenic co-occurring with 'pills' should be much higher than its co-occurrence with 'women' (and similarly for the other combination). Another important point Levesque makes in proposing the WS challenge is avoiding posing queries that are either too obvious, or too difficult. The latter could happen, for example, if the questions posed required knowledge of a special vocabulary that only specialized domain experts might know. Essentially, good WS sentences should be ones that a 5-year old would be able to effortlessly answer – or, as Levesque puts it, "a good question for a WS is one that an untrained subject (your Aunt Edna, say) can answer immediately". The question of what makes a good WS sentence is thus crucial. In the next section we suggest how this question can be more formally and systematically answered.

## Situating the Winograd Schema in the Data-Information-Knowledge Continuum

To systematically deal with Levesque's concern of not posing questions that are too easy or too difficult we consider such questions at a higher-level of abstraction. All WS queries share the following features:

a) there are two noun phrases in a WS sentence
b) there is a reference made to one of these noun phrases
c) the question involves determining the right referent

This template applies to four scenarios that correspond to the *distance* of the relevant information needed to resolve the reference (how far is the relevant information from the surface data). In general, the reference (*i*) can be resolved by simple lexical and/or **syntactic** information available in the (data of the) sentence itself (Level 1); (*ii*) can be resolved by **semantic** information in the form of properties or attributes of some of the data in the sentence (Level 2); (*iii*) can only be resolved by reasoning at the **pragmatic** level by accessing commonsense background knowledge about the entities and relations mentioned in the sentence (Level 3); or (*iv*) cannot be resolved at all unless the overall intent and discourse is brought to light (Level 4) (see Figure 1). We will not be concerned here with sentences at Level 4, where the reference cannot be resolved unless additional discourse-level information is brought to light, as is the case in the sentence '***Jon** told **Bill** that **he** wasn't selected by the committee*'. Below we discuss Levels 1 through Level 3 in some detail.

**The Syntactic/Data Level**
This is the level at which only one of the two noun phrases is the correct referent and where simple lexical/syntactic information available in the data is enough to resolve the reference. Here are two typical examples:

(5) a. ***John*** *informed **Mary** that **he** passed the exam.*
   b. ***John*** *invited **his classmates** for his birthday party, and most of **them** showed up.*

The references in (5a) and (5b) can be easily resolved using information that is readily available as *attributes* of the lexical data - in particular, the references in (5) can be resolved by ensuring `gender` (male/female/neutral) and `number` (singular/plural) agreement.

**The Semantic/Information Level**
At this level the information required is not readily available as *attributes* of the lexical data, but is one step away, in the form of *relations* between the various lexical items in the data. (6) is a typical example illustrating this situation:

(6) *Our **graduate students** published* 20 ***papers** this year and, apparently, few of them*
   a. ***authored*** *some books*
   b. ***appeared*** *in top journals*

The reference in (6) can be easily resolved using the type constraints (or 'selectional restrictions') AUTHOR(human, content) and APPEARIN(content, publication) enforcing the following: humans, and not papers, author content,

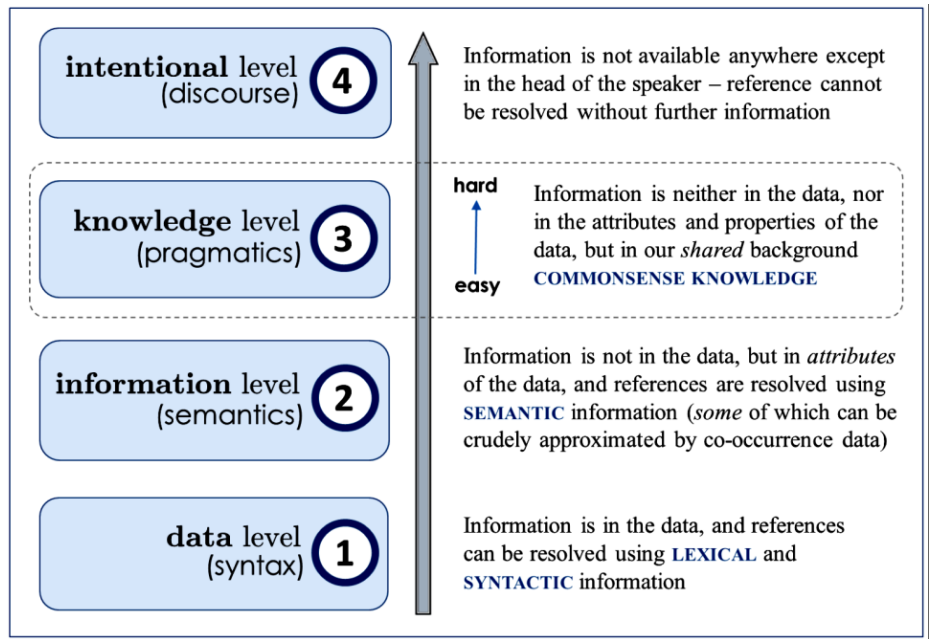

**Figure** 1. The Winograd Schema situated in the data-information-knowledge continuum

and it is content, and not students, that appear in a publication. What should be noted at this point is that, once the relevant information becomes available, resolving the reference in both levels 1 and 2 is certain: the plausibility of all, but one referent only, becomes 0. For example, in (5a) and (6a) we have the following, respectively:

$P(he=John) = P(\mathbf{gender}(he) = \mathbf{gender}(John)) = 1$
$P(he=Mary) = P(\mathbf{gender}(he) = \mathbf{gender}(Mary)) = 0$
$\therefore P(he=John) > P(he=Mary)$

$P(them=students) = P(\text{AUTHOR}(student, book)) > 0$
$P(them= papers) = P(\text{AUTHOR}(paper, book)) = 0$
$\therefore P(them=students) > P(them= papers)$

The point here is that once the relevant information becomes available (whether immediately from the sentence data, as in Level 1 or using information one step away from the surface data, as in Level 2), the resolution of the reference in levels 1 and 2 is certain - one referent becomes valid while all others will not even be 'possible' candidates. The situation is quite different at the pragmatic level (Level 3), however, as discussed below.

### The Pragmatic/Knowledge Level

This is the level at which 'good' WS sentences are situated. Sentences at this level are those where the reference in question can, in theory, be resolved by either of the two noun phrases, and where the 'most appropriate' referent is usually the one that is more **plausible** among all the **possible** candidates, and where *the more plausible referent is the one that makes the final scenario being described more compatible with our commonsense understanding of the world*. Care should therefore be exercised in **not** choosing WS sentences where the *likelihood* of both referents are near equal (these are the cases where the WS is too difficult), or where the likelihood of one is clearly much higher than the other (cases where the WS is too easy). Shown in table 1 below are examples that illustrate WS sentences at the pragmatic (knowledge) level.

There are several important things to note here: (*i*) unlike the situation in Level 1 and Level 2, referents in the examples of table 1 are both, in theory, **possible** (i.e., $P(referent1) > 0$ and $P(referent2) > 0$) although the plausibility of one is higher than the plausibility of the other - i.e. $\mathbf{P}(referent1) > \mathbf{P}(referent2)$ or $\mathbf{P}(referent2) > \mathbf{P}(referent1)$[3]; and (*ii*) unlike the situation in levels 1 and 2, the references in table 1 cannot be resolved by simple attributes (or relations between attributes) of the lexical items in the sentence data, but requires some background knowledge. For example, if SHOT($x$, $y$) holds between some $x$ and some $y$, then it is more likely for $x$ to try to escape and more likely for $y$ to try to arrest $x$. Similarly, if ~LIFT($x$, $y$) is true - that is, if $x$ cannot lift $y$, then TOO-HEAVY($y$) is more likely than TOO-HEAVY($x$), and if ~FIT($x$,$y$) then TOO-SMALL($y$) is more likely than TOO-SMALL($x$), and TOO-BIG($x$) is more likely than TOO-BIG($y$), etc.

---

[3] While $P(x)$ is the probability of $x$, we use $\mathbf{P}(x)$ to refer to the *plausibility* of $x$ which (for our purposes) is the degree to which $x$ is compatible with our commonsense view of the world (the exact nature of $\mathbf{P}$ in the context of language understanding is of course an interesting topic on its own).

| | |
|---|---|
| ***A teenager** shot **a policeman** but eyewitnesses said he managed to*<br>a. *escape*<br>b. *arrest him* | SHOT($x, y$) ⊃ **P**(ESCAPE($x$)) > **P**(ESCAPE($y$))<br>∧ **P**(ARREST($y, x$)) > **P**(ARREST($x, y$)) |
| ***John** could not lift his **son** because he was too*<br>a. *heavy*<br>b. *weak* | ~ LIFT($x, y$) ⊃ **P**(TOO-HEAVY($y$)) > **P**(TOO-HEAVY($x$))<br>∧ **P**(TOO-WEAK($x$)) > **P**(TOO-WEAK($y$)) |
| *The **trophy** would not fit into **the brown suitcase** because it was too*<br>a. *big*<br>b. *small* | ~ FIT($x, y$) ⊃ **P**(TOO-SMALL($y$)) > **P**(TOO-SMALL($x$))<br>∧ **P**(TOO-BIG($x$)) > **P**(TOO-BIG($y$)) |
| *The **town councilors** refused to give the angry **demonstrators** a permit because they*<br>a. *feared violence*<br>b. *advocated violence* | ANGRY-DEMONSTRATORS($x$) ∧ CITY-COUNCILORS($y$)<br>⊃ **P**(FEAR-VIOLENCE($y$)) > **P**(FEAR-VIOLENCE($x$))<br>∧ **P**(ADVOCATE-VIOLENCE($y$)) > **P**(ADVOCATE-VIOLENCE($x$)) |

**Table 1.** WS sentences and the plausibility of the two (equally possible) referents.

To summarize, ideal WS sentences are those where both referents are in theory possible, and where the information required to resolve the reference cannot be obtained from syntactic data nor from semantic information, but is obtained from background knowledge that, once available makes one of the **possible** referents more **plausible**.

# The 'Missing Text Phenomenon': is the Winograd Schema just a Special Case?

Having discussed the Winograd Schema (WS) in some detail, suggesting in the process where good WS sentences are situated, we would like to suggest here that WS sentences are in fact special cases of a more general phenomena in natural language understanding that a good test for machine intelligence must also consider.

The sentences at Level 3 are good WS sentences specifically because these are typical examples where the challenge is to *infer* the missing text - text ***that is not explicitly stated in the text*** but is assumed as shared commonsense knowledge. As Levesque (2012), noted:

> "You need to have background knowledge that is not expressed in the words of the sentence to be able to sort out what is going on …. And it is precisely bringing this background knowledge to bear that we informally call *thinking*." (Emphasis added)

We wholeheartedly agree: what we call *thinking* in the process of language understanding is precisely that ability to determine the most **plausible** scenario among all **possible** scenarios, and this is done by having access to information that is not explicitly stated in the text but is assumed among a speakers of ordinary language as commonsense (background) knowledge. However, this 'missing text phenomenon' (which we will refer to as MTP), of accessing background knowledge not explicitly stated in the text, is not specific to reference resolution, but is in fact the common denominator in many other linguistic phenomena. Below we briefly discuss how the MTP is the source of semantic challenges involving a number of linguistic phenomena other than reference resolution.

## MTP and Hidden Events in Relational Nominals

Consider the examples in (7) (Pustejovsky et. al., 1988):

(7)  a. *John enjoyed* **[reading]** *the book*
     b. *John enjoyed* **[watching]** *the movie*
     c. *John enjoyed* **[smoking]** *the cigarette.*

While *John* can, in theory, enjoy writing, publishing, buying, or selling a book, and enjoy directing, producing, buying, selling, a movie, etc., a 5-year old would immediately infer the **[missing text]** in (7) and precisely because the most plausible hidden verb is the one that is *more consistent with our commonsense understanding of the world*: the most salient relation between people and books is 'reading', that between people and movies is 'watching', etc. If such examples were to be part of the WS challenge then a query posed against such sentences would be "what did John enjoy about the book" for (7a) and "what did John enjoy about the movie?" for (7b), where the answers to choose from could be two or more 'possible' answers (reading/selling/buying, etc.).

## MTP and Prepositional Phrase Attachments

Consider the sentence pairs in (8) which are examples of *prepositional phrase (PP) attachments*.

(8)  *I read a story about evolution in the last ten*
     a. *minutes.*
     b. *million years.*

Clearly, the most plausible interpretation of (8a) is '*I read a story about evolution* **[and finished it]** *in the last ten minutes*' while the correct interpretation of (8b) is '*I read a story about evolution* **[that occurred]** *in the last ten mil-* lion years'. Again, the ambiguity is due to the 'missing text' that can only be uncovered using background commonsense knowledge: (*i*) evolution cannot happen in 10 minutes, but the act of reading a story could; and (*ii*) our commonsense understanding of the world precludes the reading of a story to take 10 million years. If such sentences were to be used in the WS challenge, then a good question to (8a) and (8b) would be: what is that took ten minutes/million years (answers: evolution/reading)?

## MTP and Quantifier Scope Ambiguities

In (9) we have an example where we need to resolve what is referred to in the literature as *quantifier scope ambiguities* by, again, accessing the relevant commonsense background knowledge to infer the **[missing text]** that is not usually explicitly stated.

(9) *John visited a* **[different]** *house on every street in his neighborhood.*

Inferring the missing text is what allows us here to reverse the surface scope ordering and interpret (9) as '*On every street in his neighborhood, John visited a house*'. If such questions were to be used in the WS challenge, then a good question for (9) would be: how many houses does (9) refer to (and the answers could be 1 and many)

## MTP and Metonymy

What is referred to in the literature as metonymy is yet another example of where humans use commonsense background knowledge to infer the **[missing text]**, as illustrated by the sentences in (10).

(10) a. *The omelet wants another beer.*
     ⇒ *The* **[person eating the]** *omelet wants another beer.*
  b. *The car in front of us is annoying me, pass it please.*
     ⇒ *The* **[person driving the]** *car in front us is annoying me, pass it please.*

For such sentences to be part of the WS challenge, a question such as this can posed for the sentence in (10a): 'what is the *type* of object that wants a beer?' And the alternative answers would be person/table.

The main point of this section was to illustrate that, besides reference resolution, what we usually call *thinking* in the process of language understanding almost always involves discovering a significant amount of missing text, text that is not explicitly stated but is assumed as shared background knowledge. The crucial question now is this: if the missing text phenomenon (MTP) is correct - i.e., if in our ordinary discourse we (in accordance with one of Grice's maxims to make our statements as informative as required **but not more**) tend to leave out what we assume is shared background knowledge, then how could there be a viable data-driven approach to language understanding? We discuss this next.

## Data-Driven Language Understanding?

In this section we will suggest that the 'missing text phenomenon' (MTP) places severe limitations on the data-driven and machine learning approaches to natural language understanding. The first argument is a technical one, and it is based on theoretical results where the equivalence between learnability and compressibility has been established - see, for example (David et. al., 2016) and the more recent (Ben-David, et. al. 2019). Essentially, what these results tell us is that learnability can only occur if the data we are learning from is compressible (and vice versa). However, and as argued above at length, much of what we call *thinking* in the process of language understanding is about discovering the 'missing text' (the text we leave out), and thus ordinary spoken language is not compressible as it is already highly (and optimally!) compressed. And given the equivalence of learnability and compressibility, thus, ordinary spoken language cannot be learned[4]. What's at issue here is this: while the data-driven machine learning approach is an attempt at generalizing and compressing the data by finding meaningful patterns, the language understanding problem is about uncompressing - in fact, it is about expanding and amplifying, by 'uncovering' all the hidden text! It would seem therefore that the goal of machine learning and that of language understanding are at odds, to put it mildly!

But despite this (perhaps controversial) argument, the data-driven/machine learning approach to language understanding can be questioned on other grounds that are more relevant to our discussion of the WS challenge. Consider again the sentence in (11), discussed above in Table 1.

(11) *The* **trophy** *would not fit into* **the brown suitcase** *because it was too*
   a. **big**
   b. **small**

The most obvious way to learn how to resolve the reference in (10) using a purely data-driven/machine learning approach would be to, essentially, and given a large corpus, find out the following probabilities:

$p_{11}$ = P(*The trophy… because* [**the trophy**] *was too small*)
$p_{12}$ = P(*The trophy…because* [**the suitcase**] *was too small*)
$p_{21}$ = P(*The trophy … because* [**the trophy**] *was too big*)
$p_{22}$ = P(*The trophy ... because* [**the suitcase**] *was too big*)

That is, a machine learning approach to resolving such references would essentially try to find out which replacement is more likely to be found in a large corpus. This general approach has indeed been tried by (Trinh and Le, 2018). As pointed out in (Saba, 2018b), however, such an approach will not scale as the replacement of the reference 'it' by one of the referents and computing the probability of each replacement in a large corpus is not enough. For example, the preferred referent would also change if 'would not fit' was replaced in (11) by 'would fit', or 'because' was changed to 'although', etc. Moreover, and since data-driven and machine learning approaches do not admit an ontological structure where objects similar to 'trophy' and objects similar to 'suitcase' are arranged in a typed-hierarchy, another set of probabilities would have to be computed for sentences where 'trophy' is replaced by 'laptop', for example, and where 'suitcase' was replaced by 'bag', etc. Simple calculations would show that a data-driven approach would need to process hundreds of millions of examples, just to learn how to resolve references in sentences similar to those in (11). What seems to be happening here is that a data-driven approach to language understanding would need to replace the 'uncovering' of the missing data by a futile attempt at memorizing most of language - something that is theoretically, not to mention computationally and psychologically, implausible.

---

[4] It should be noted here that what we are questioning is the data-driven and machine learning approach to natural language **understanding**; meaning that specific task of building systems that can comprehend ordinary spoken language, much like humans do, and not other 'text processing' tasks such as document classification, text clustering, spam filtering, etc. where the text in question can be treated like any other data, and where the text can, in theory, be compressed.

## Concluding Remarks

In this paper we suggested where appropriate WS sentences are situated in the data-information-knowledge continuum. In particular, we suggested that 'good' WS sentences are those that cannot be answered using syntactic **data** or semantic **information**, but can only be resolved at the pragmatic level by uncovering the missing text - text that is never explicitly stated but is assumed as shared background **knowledge**. We further suggested that this 'missing text phenomenon' (MTP) is not specific to reference resolution but to most challenges in the semantics of natural language and suggested further how the WS can be extended to include such linguistic phenomena. Against the backdrop of MTP we further argued that this phenomenon precludes data-driven and machine learning approaches from providing any real insights into the general problem of natural language understanding.

Another aspect of this work that could not be discussed here for lack of space is related to why ignoring MTP is perhaps the reason logical semantics might have faltered. To see the relation of MTP to problems in traditional logical semantics, consider (12a) and (12b).

(12) a. *Julie is an articulate person*
$\Rightarrow$ ARTICULATE(*Julie*) $\land$ PERSON(*Julie*)
b. *Julie is articulate* $\Rightarrow$ ARTICULATE(*Julie*)

(12a) and (12b) have different translations into first-order predicate logic, although the two sentences seem to have the same semantic and cognitive content. One way to resolve this semantic puzzle is to acknowledge the difference between ontological concepts - that are types in a strongly-typed ontology, and logical concepts - that are the properties of and the relations between various ontological types. As such, the proper translation of (12a) and (12b) would be to assume that, in the context of being ARTICULATE, PERSON(*Julie*) is true, a priori - in other words, that the property ARTICULATE is said of objects that are of type **person**:

(13)    ($\exists$*Julie* :: **person**)(ARTICULATE(*Julie*))

Embedding ontological types in our semantics allows us then to uncover all the missing text as for example in (14).

(14) *The omelette wants another beer*
$\Rightarrow$ ($\exists o$ :: **omelette**)($\exists b$ :: **beer**)
    (WANT($o$ :: **person**, $b$ :: **entity**))

Note now that the 'hidden text' (namely, that it is the person eating the omelette who wants the beer), can be uncovered by unifying **omelette** with the expected types of WANT. Details of this work, that attempts to rectify a major oversight in logical semantics, namely how MTP was completely ignored, can be found in (Saba, 2018a).